\documentclass[fleqn,10pt]{wlscirep}
\usepackage{amsmath,amsfonts}
\usepackage{algorithmic}
\usepackage{algorithm}
\usepackage{array}
\usepackage[utf8]{inputenc}
\usepackage[T1]{fontenc}
\usepackage{textcomp}
\usepackage{stfloats}
\usepackage{url}
\usepackage{verbatim}
\usepackage{graphicx}
\usepackage{cite}
\usepackage{amsmath,amssymb,amsfonts}
\usepackage{soul}
\usepackage{algorithm}
\usepackage{hyperref}
\usepackage{algorithmic}
\title{Towards Knowledge-Infused Automated Disease Diagnosis Assistant}  

\author[1,*]{Mohit Tomar}
\author[1,*]{Abhisek Tiwari}
\author[1, +]{Sriparna Saha}
\affil[1]{Department of Computer Science and Engineering, Indian Institute of Technology, Patna, 801103, India}
\affil[*]{These authors contributed equally to this work}
\affil[+]{Corresponding author}

\begin{abstract}
With the advancement of internet communication and telemedicine, people are increasingly turning to the web for various healthcare activities. With an ever-increasing number of diseases and symptoms, diagnosing patients becomes challenging. In this work, we build a diagnosis assistant to assist doctors, which identifies diseases based on patient-doctor interaction. During diagnosis, doctors utilize both symptomatology knowledge and diagnostic experience to identify diseases accurately and efficiently. Inspired by this, we investigate the role of medical knowledge in disease diagnosis through doctor-patient interaction. We propose a two-channel, knowledge-infused, discourse-aware disease diagnosis model ({\em KI-DDI}), where the first channel encodes patient-doctor communication using a transformer-based encoder, while the other creates an embedding of symptom-disease using a graph attention network (GAT). In the next stage, the conversation and knowledge graph embeddings are infused together and fed to a deep neural network for disease identification. Furthermore, we first develop an empathetic conversational medical corpus comprising conversations between patients and doctors, annotated with intent and symptoms information. The proposed model demonstrates a significant improvement over the existing state-of-the-art models, establishing the crucial roles of (a) a doctor's effort for additional symptom extraction (in addition to patient self-report) and (b) infusing medical knowledge in identifying diseases effectively. Many times, patients also show their medical conditions, which acts as crucial evidence in diagnosis. Therefore, integrating visual sensory information would represent an effective avenue for enhancing the capabilities of diagnostic assistants.
\end{abstract}
\begin{document}

\flushbottom
\maketitle
\thispagestyle{empty}

% \noindent Please note: Abbreviations should be introduced at the first mention in the main text – no abbreviations lists. Suggested structure of main text (not enforced) is provided below.

\section*{Introduction}
The development of the Internet was primarily aimed at providing global access to information. In the last few years, the Internet has become one of the most popular and reliable platforms for accessing healthcare-related information. A survey by Cohen et al. \cite{cohen2011use} found that more than 65\% of US adults use the Internet for performing several healthcare-related activities. Over the past five years, numerous surveys have highlighted an alarming population-to-doctor ratio in different countries, emphasizing the urgent need for improvements in healthcare systems. According to the report of the World Health Organisation (WHO), 2013 \cite{george2014online}, there is a shortage of 7.2 million health workers globally which can reach 12.9 million in the upcoming decade. With the motivation of assisting doctors and utilizing their time more efficiently, there has been a significant rise in the popularity of artificial intelligence-based virtual assistants and tools for various medical activities, including automatic disease diagnosis. The objective of Automatic Disease Diagnosis (ADD) \cite{wei2018task,teixeira2021interplay,liao2020task,peng2018refuel} is to support doctors by performing an initial examination of symptoms. It also diagnoses disease from the conversation between the patient and the doctor. First, the user reports their problems and symptoms (called explicit symptoms) in their self-report, and then the agent inquires about additional symptoms (called implicit symptoms) to diagnose the disease. Hence, an automatic disease diagnosis system can be summarised as a system where an agent inquires about symptoms step by step and then can diagnose disease based on implicit and explicit symptoms.
Hence, in a healthcare setting that incorporates this system, when a patient visits a doctor, the doctor is provided with comprehensive information about the patient and his/her situation. Some automatic disease diagnoses systems, such as Mayo Clinic, Babylon Healthcare, and GMAN \cite{yuan2021graph}, are already deployed, which are being extensively used by both hospitals and end-users. 

\begin{figure}[hbt!]
    \centering
    \includegraphics[scale=0.37]{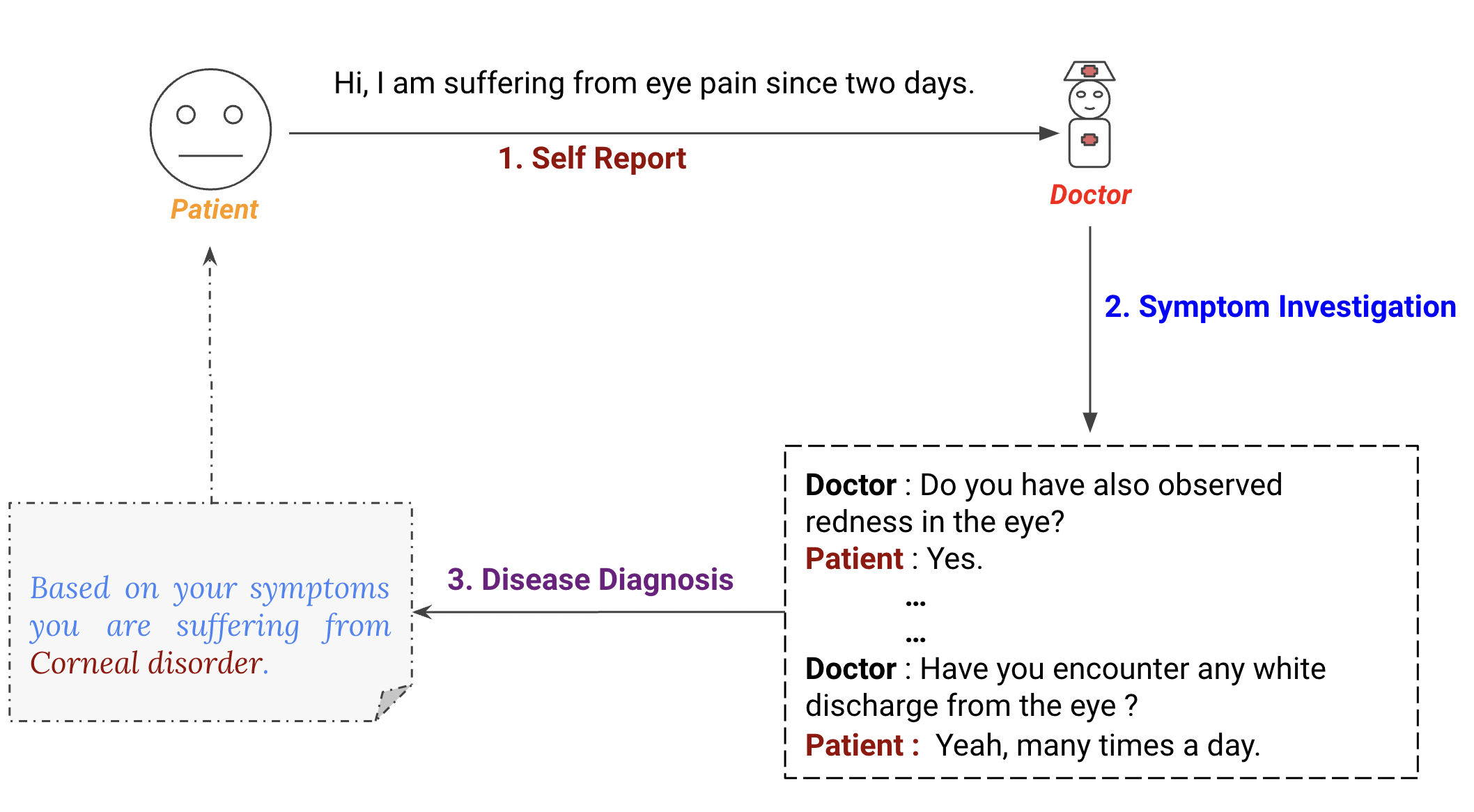}
    % \vspace{-2.5em}
    \caption{An illustration of online symptom investigation and disease diagnosis}
    \label{f1}
\end{figure}

In online communication with doctors, patients first inform their chief complaints, known as self-reports to doctors. Based on the chief complaint, a doctor is assigned, who conducts a detailed symptom investigation and extracts relevant symptoms through chat. An example is shown in Figure \ref{f1}. Over the last five years, significant efforts have been made by both the dialogue and healthcare communities to develop an artificial intelligence-based diagnosis assistant that can act as a third eye for disease diagnosis \cite{yu2021reinforcement, kumar2022artificial}. In the study \cite{wei2018task}, they introduced a task-oriented dialogue system, which collects patient self-reported information and extracts further signs and symptoms during conversational interactions. In \cite{peng2018refuel}, the authors have illustrated the impact of different reward functions utilized to provide feedback to a reinforcement learning-based diagnosis system on diagnosis efficacy. Following the work, the work \cite{liao2020task} incorporated a medical department-driven disease diagnosis system, which illustrated superior performance in terms of both quantitative and qualitative metrics. However, most of the diagnosis assistants \cite{liao2020task, kao2018context, liao2020task} are based on some data-driven approaches, which learn solely from existing and underlying medical corpora. Thus, given the scarcity of publicly available medical corpora, they are likely to result in a model with local knowledge concentrated in the underlying corpus. In real life, doctors also learn from knowledge bases and well-established principles in addition to diagnosis experience.

 It is common for us to communicate only our most prominent and urgent health concerns to doctors during consultations. However, doctors do not rely solely on our reported symptoms to make diagnoses and prescribe treatments. Instead, they conduct a thorough investigation to arrive at a conclusive diagnosis. This is necessary because we tend to report only the most common and noticeable symptoms, overlooking other potential clues. They collect additional evidence to better understand medical issues and treat them properly. Motivated by the above two observations, we aim to build knowledge-infused, discourse-aware disease identification (KI-DDI) that incorporates an external knowledge graph and uses an attention mechanism that emphasizes the importance of self-report in the whole conversation. We also create a symptom disease knowledge graph (S-S-D) where symptoms and diseases act as nodes, and an edge between them is treated as a co-occurrence of both of them. We then determine edge weights using the symptom frequency-inverse disease frequency (sf-idf) method, inspired by the term frequency-inverse document frequency (tf-idf) method \cite{ramos2003using}. The edge weight between a symptom and a disease determines the co-occurrence of the symptom and disease. In KI-DDI, we pass the whole dialog to the language model to extract embedding. We then extract symptoms from the dialog and retrieve the sub-graph from the knowledge graph relevant to the dialog. We then form a joint graph by connecting the dialog node and subgraph. Finally, we obtain graph embedding by considering the mean pool, and then an attention mechanism is used to calculate the weighted sum of the dialog node and self-report, where graph embedding is infused with dialogue embedding to perform disease classification.

In the last few years, there have been tremendous efforts by both research and industrial communities to automatize many medical operations to assist physicians \cite{davenport2019potential}. Nevertheless, the exploration and outcome of these efforts are limited primarily due to the lack of an adequate amount of medical data \cite{miotto2018deep}. For example, there is not a single conversational disease diagnosis dialogue corpus in English. Motivated by the limitation, we curate an empathetic medical dialogue dataset called Empathical Dialogue Dataset. We annotate each utterance of a conversation with its corresponding intent and symptom information. There are two types of intent tags: Symptom and Affirmative. Symptom intent indicates the presence of a symptom, and Affirmative intent indicates that the patient agrees with the doctor, but the symptom is not present in the patient's utterance. The role of empathy in this dataset is that it helps patients feel trusted and cared for by the doctor. The dialogue corpus bridges the following gaps in the medical diagnosis research community: (a) End-to-End communications directly with end-users in the English language, (b) Medical utterance understanding modules could be pre-trained using the curated corpus for symptom extraction, and (c) Context coherent response generation. 

\hspace{-0.55cm}\textbf{Research Questions and Hypotheses} In this paper, we investigate the following three research questions: \textbf{(i)}  Are self-reports from patients sufficient for an accurate diagnosis? \textit{We hypothesize that the patient’s self-report (first utterance by the patient) alone is insufficient for disease diagnosis. We performed empirical studies that showed that self-report is insufficient in diagnosing disease as the model achieves poor diagnosis accuracy. Thus, it indicates the need for further symptom investigation}.
\textbf{(ii)}  How does the medical knowledge graph influence the disease diagnosis model’s performance? {\textit{We hypothesize that the external medical knowledge graph aids in disease diagnosis. We showed through empirical studies that our model KI-DDI incorporating external knowledge outperformed the baseline models without external knowledge. This shows external knowledge provides valuable insights in diagnosing disease.
}}
\textbf{(iii)} Does the mechanism of knowledge infusion impact the efficacy of disease diagnosis? {\textit{We hypothesize that incorporating knowledge using graph structure is an efficient mechanism of infusing them. We showed through empirical studies that when external knowledge is infused in a graph structure, it performs better than when it is infused in a linear structure. Thus, adding external knowledge in a graph structure helps diagnose disease.
}}

\hspace{-0.55cm}\textbf{ Key Contributions} The key contributions of the work are three-fold, which are enumerated as follows: 
\vspace{-0.5em}
\begin{itemize}
    \item  We propose a two-channeled knowledge-infused, discourse-aware disease identification (KI-DDI) model that leverages external medical knowledge encoded through a context-aware filtered knowledge graph for identifying diseases accurately and efficiently from patient-doctor communications.  
    \vspace{-0.5em}
    \item We first curate a conversational medical dialogue corpus named Empathical dialogue dataset in English, where each utterance is annotated with its corresponding intent and slot information.
    \vspace{-0.5em}
    \item The proposed KI-DDI model achieves a significant improvement over an existing state-of-the-art model and establishes a new benchmark for the conversation-driven diagnosis problem. 
\end{itemize}

% \hspace{-0.55cm}\textbf{Reproducibility:}

\section*{Background}

\subsection*{Related Works}

The research primarily pertains to the following areas: electronic health records, automatic disease diagnosis, graph neural network, knowledge infusion, and Dynamic Uncertain Causality Graph. In the subsequent paragraphs, we provide summaries of the pertinent works in these domains. 
% and highlighted the research gaps. 

\hspace{-0.55cm}\textbf{Electronic health records (EHR)} During the early 2000s, systems based on Electronic Health Records (EHR) were introduced with the goal of aiding patients, driven by the motivation to provide virtual assistance to individuals in rural areas\cite{ventres2006physicians} In BEHRT \cite{li2020behrt}, the authors developed a transformer-based model for mining electronic health records (EHR). It uses patients' EHR data to perform multi-label classification for given all possible diseases. It is also capable of personalized recommendations, and it can incorporate concepts such as diagnosis, medication, and measurements. In \cite{li2022electronic} authors proposed a reinforcement learning algorithm based on EHR to optimize the sequential processing of diseases. It considers both physiological variables and major disease factors during EHR
modeling to improve the interpretability of the model. It utilizes Deep Q Learning (DQN) \cite{mnih2013playing} algorithm to explore the optimal insulin dosage for the patients. In \cite{nemesure2021predictive}, authors handled the problem of Generalized Anxiety Disorder (GAD) and Major Depressive Disorder (MAD) using an ensemble of machine learning pipelines (Support Vector Machine, XG Boost, K Nearest Neighbor, Random Forest, Logistic Regression, Neural Network). It also utilized SHAP values to highlight which features had the major impact on the prediction for each disease. In Med-BERT \cite{rasmy2021med}, authors adapt BERT \cite{devlin2018bert} in the EHR setting. As an input, it receives three types of embeddings: the diagnosis code, the order of code within each visit, and the position of each visit. It achieved remarkable performance when fine-tuned on EHR. In Med7 \cite{kormilitzin2021med7}, authors introduced a named-entity recognition model that is trained on EHR. The goal of the model is to recognize seven categories such as drug names, route of administration, frequency, dosage, strength, form, and duration.

\hspace{-0.55cm}\textbf{Automatic disease diagnosis} The utilization of an Electronic Health Record (EHR) system necessitates the coordination and synchronization of multiple devices \cite{menachemi2011benefits}. 
To streamline the process, researchers have introduced a novel technique for automatically diagnosing non-fatal or sensitive diseases. In this approach, an interactive system conducts symptom investigation and provides disease diagnoses\cite{tang2016inquire}. Wei et al. \cite{wei2018task} devised a task-oriented dialogue procedure for symptom investigation.
In this process, the agent gathers symptoms through conversation and subsequently diagnoses a disease based on the observed symptoms.
In \cite{kao2018context}, the authors presented a context-aware symptom checker, which also models patients' personal information, such as gender and age, in disease diagnosis. The experimental results of the contextual model confirm the vital role of patients' personal information in executing an appropriate and efficient diagnosis. 
Liao et al. (Liao 2020 Task) have proposed an integrated and synchronized two-level policy framework using hierarchical reinforcement learning \cite{dietterich2000hierarchical}.
The model demonstrated superior performance compared to the flat policy approach \cite{wei2018task} by a considerable margin. In \cite{chen2022diaformer}, authors considered disease diagnosis as a generation process. It uses a symptom attention framework for the generation of symptoms and diagnosis. It uses an orderless training mechanism. 

\hspace{-0.55cm}\textbf{Graph Neural Network} Graph Convolutional Network (GCN) \cite{kipf2016semi} uses graph data and updates the node embedding depending upon the neighboring nodes. Graph Attention Network (GAT) \cite {velivckovic2017graph} uses an attention mechanism to get the embedding of nodes depending on which neighboring node is relevant. In Graph Transformer \cite{Dwivedi2020AGO}, authors proposed the adaptation of the transformer network to graphs. It uses an attention mechanism that depends on a neighboring connection for each node. In \cite{rampavsek2022recipe}, authors present a systematic approach to building a scalable graph transformer. Its time complexity is linear in the number of nodes and edges. In \cite{zhu2021simple}, authors used a modified Markov decision kernel to derive Simple Spectral Graph Convolution. Upon using this method, it trades off between low and high pass filter bands, which capture global and local context for each node. In \cite{li2021training}, authors found that using reversible connections with deep networks allows the effective training of an overparameterized graph neural network. In \cite{brody2021attentive}, authors showed the limitation of using static attention in a graph attention network (GAT). They further developed the dynamic attention mechanism, which attends dynamically to neighboring nodes depending on the query node, to overcome the limitation.

\hspace{-0.55cm}\textbf{Knowledge Graph and Knowledge Infusion} 
Numerous studies have been conducted to integrate external knowledge into the language model. ERNIE \cite{zhang2019ernie} adds external knowledge by infusion of token embedding and entity embedding. It has an information fusion layer that mixes token embedding and its corresponding entity embedding. In \cite{yasunaga2021qa}, it retrieves the subgraph based on the entities. Further, it forms the joint graph of language embedding and subgraph and applies Graph Neural Network (GNN) for knowledge infusion. GreaseLM \cite{zhang2022greaselm} focuses on the deep fusion of embeddings from the language model and the graph neural network using a modality interaction unit over multiple layers. In \cite{yasunaga2022deep} authors utilized self-supervised learning methods such as masked language modeling and knowledge graph link prediction for learning joint representation of text and knowledge graph. 
In \cite{milewski2022finding} authors studied the capabilities of the multimodal BERT model in storing the grammatical and linguistic knowledge that is learned with the help of objects in images. In \cite{liu2021generated}, authors developed the method of knowledge prompting in which they first extracted knowledge from a language model, and later they used that knowledge in question-answering tasks. 

\hspace{-0.55cm}\textbf{Dynamic Uncertain Causality Graph (DUCG)} has been used for the purpose of the clinical diagnosis. DUCG\cite{dong2014methodology} has been utilized to diagnose vertigo by incorporating symptoms, signs, medical histories, etiology, and pathogenesis. Also, Cubic DUCG\cite{8949696} has been used for fault diagnosis for complex systems by representing dynamic casualties in the system fault spreading process in a compact manner and conducting accurate reasoning.
% For clinical diagnosis\cite{zhang2021dynamic}, DUCG is being utilized by solving problems of influences of risk factors, simultaneous consequential observations, treatment of isolated state-abnormal observations, and disease-specific manifestations. 
Also, in the context of diagnosing and treating Hepatitis B., DUCG\cite{deng2021application} based diagnosis and Treatment Unification Model is utilized. It uses Reverse logic gates to enhance the accuracy of treatment planning.

\subsection*{Problem Formulation}
The proposed model aims to identify the disease of the patient based on patient-doctor interaction. Thus, the input to an autonomous system will be dialogue, and the output will be disease. A dialogue can be regarded as sequences of patient and doctor utterances, i.e., $D = <(P_1, D_1) (P_2, D_2) ...... (P_n, D_n) >$ where $(P_i, D_i)$  denote $i^{th}$ utterance of patient and doctor, respectively, and $n$ signifies the total number of turns in the dialogue. The disease identification through patient-doctor dialogue can be expressed as follows:

\begin{equation}
\small
    d = argmax_j P (Dis_j \vert \{(P_1, D_1) (P_2, D_2) ...... (P_n, D_n)\}, \theta)
\end{equation}
where $Dis$ is the set of diseases, the term, $\theta$ denotes the diagnosis model's parameter. 

\section*{Dataset}
% \section*{Dataset.} 
 We begin by investigating the benchmark medical diagnosis dialogue datasets, and the findings are presented in Table \ref{ED}. We could not find a single dyadic conversational diagnosis dataset in English, which motivated us to curate a new medical dialogue corpus. Doctors usually engage with and respond empathetically to their patients, which increases patient compliance and further helps in building trust between patient and doctor. We developed an Empathetic Medical (Empathical) Dialogue dataset with the help of the benchmarked SD \cite{zhong2022hierarchical} dataset and clinical guidelines provided by medical experts. 

\begin{table*}[hbt!]
    \centering
    \scalebox{1}{
    \begin{tabular}{|p{4cm}|c|c|c|c|c|c|}
    \hline
      \textbf{Dataset} & \textbf{Language}  & \textbf{Conversation} & \textbf{Intent} & \textbf{Symptom}    \\ \hline
      RD \cite{wei2018task}   &  Chinese & $\times$ & $\times$ & $\times$   \\ 
     DX \cite{xu2019end} & Chinese & $\checkmark$ & $\times$ & $\checkmark$   \\ 
     $M^2$ - MedDialogue \cite{yan2021m} & Chinese  & $\checkmark$ & $\times$ & $\checkmark$  \\
     MedDialog-EN \cite{zeng2020meddialog} & English  & $\times$ & $\times$ & $\times$   \\ 
     MedDG \cite{DBLP:journals/corr/abs-2010-07497} & Chinese  & $\checkmark$ & $\times$ & $\checkmark$   \\
     SD \cite{zhong2022hierarchical} & English & $\times$ & $\times$ & $\times$  \\ 
     Empathical (ours) & English & $\checkmark$ & $\checkmark$ & $\checkmark$  
     \\ \hline
    \end{tabular}}
    \caption{Comparison of the existing medical datasets for diagnosing disease}
    \label{ED}
\end{table*}

 \hspace{-0.63cm} \textbf{Empathical Dataset Creation and Annotation} During our investigation into the benchmarked conversation dataset, we found the SD dataset \cite{zhong2022hierarchical}, which has a database of 30K diagnosis cases covering over 90 diseases and 266 symptoms. We considered the SD dataset as a reference for creating the new conversational dataset because of its variety and credibility. We sampled 100 random diagnosis examples from the SD dataset. With the help of two clinicians, we formed a conversation-based sample dataset corresponding to the 100 diagnosis cases and annotated it with its intent and symptom information. Then we employed three medical students for the creation and annotation of dialogues based on the SD dataset samples. The students created a large dialogue corpus of 1367 diagnosis conversations by following the sample dataset and the detailed guidelines with the curated sample dataset. In order to measure the annotation agreement among the annotators, we calculated the Fleiss kappa \cite{fleiss2013statistical}, which was 0.76, indicating a strong agreement among annotators. The dataset statistics are reported in Table \ref{EMD}. 
 
 \begin{table}[hbt!]
    \centering
    \scalebox{1}{
    \begin{tabular}{|l|l|}
    \hline
     \textbf{Attribute}  &  \textbf{Value}\\ \hline
      No. of dialogues   & 1367 \\
      No. of Utterances  & 8962   \\
      Utterance tags  & {intent and symptom}\\
      Avg. dialogue length & 6.56 \\
     intent tags & symptom, affirmative \\
      No. of diseases & 90 \\ 
      No. of symptoms & 228 \\
      \hline
    \end{tabular}}
    \caption{Statistics of Empathical
 Dataset}
    \label{EMD}
\end{table}

\hspace{-0.63cm} \textbf{Clinical and Ethical Guidelines} As the medical field is highly sensitive and specialized, clinical validity holds paramount importance. We have strictly followed the guidelines established for legal, ethical, and regulatory standards in medical research during the dataset curation process. The key guidelines provided to the annotators are as follows: (i)  An annotator should not add or remove any entity in a conversation corresponding to the reported diagnosis sample in the benchmark SD dataset. (ii) No individual's personal information, which might disclose their identity, should be present in any statement within a dialogue or the entire dialogue itself. (iii) Any personal or sensitive information shared in the conversations should be properly de-identified to protect the privacy of individuals. (iv) The use of profanity or offensive language is strictly prohibited in conversations. (v) It is important to use the correct medical terminology in the transcript to ensure that the information is understood correctly by healthcare professionals. (vi) If the intent of a counseling talk is unclear, mark it, and it will be examined and confirmed by a medical professional. Furthermore, the created corpus by the annotators is thoroughly checked and verified by the clinicians. We have also obtained approval from our institute's ethical committee, IIT Patna to employ the dataset and carry out the research (IITP/EC/2022-23/07).

\hspace{-0.63cm} \textbf{Role of Dyadic Conversation and Intent/Symptom Annotation} Natural language understanding (NLU) is the first stage of a conversation system which aims to recognize users' intentions (intent) and key information from their utterances. In order to make a  disease diagnosis system that can be used for communicating directly with humans in language, NLU is necessary. Thus, we first curate the dyadic corpus and train the NLU module with the corpus. Here we have two kinds of intents (a) \textit{Symptom}, which means the presence of a symptom and (b) \textit{Affirmative} which means the patient is agreeing with a doctor, but there is no mention of the symptom in the patient's utterance.

\begin{figure*}[hbt!]
    \centering
    \includegraphics[scale=0.41]{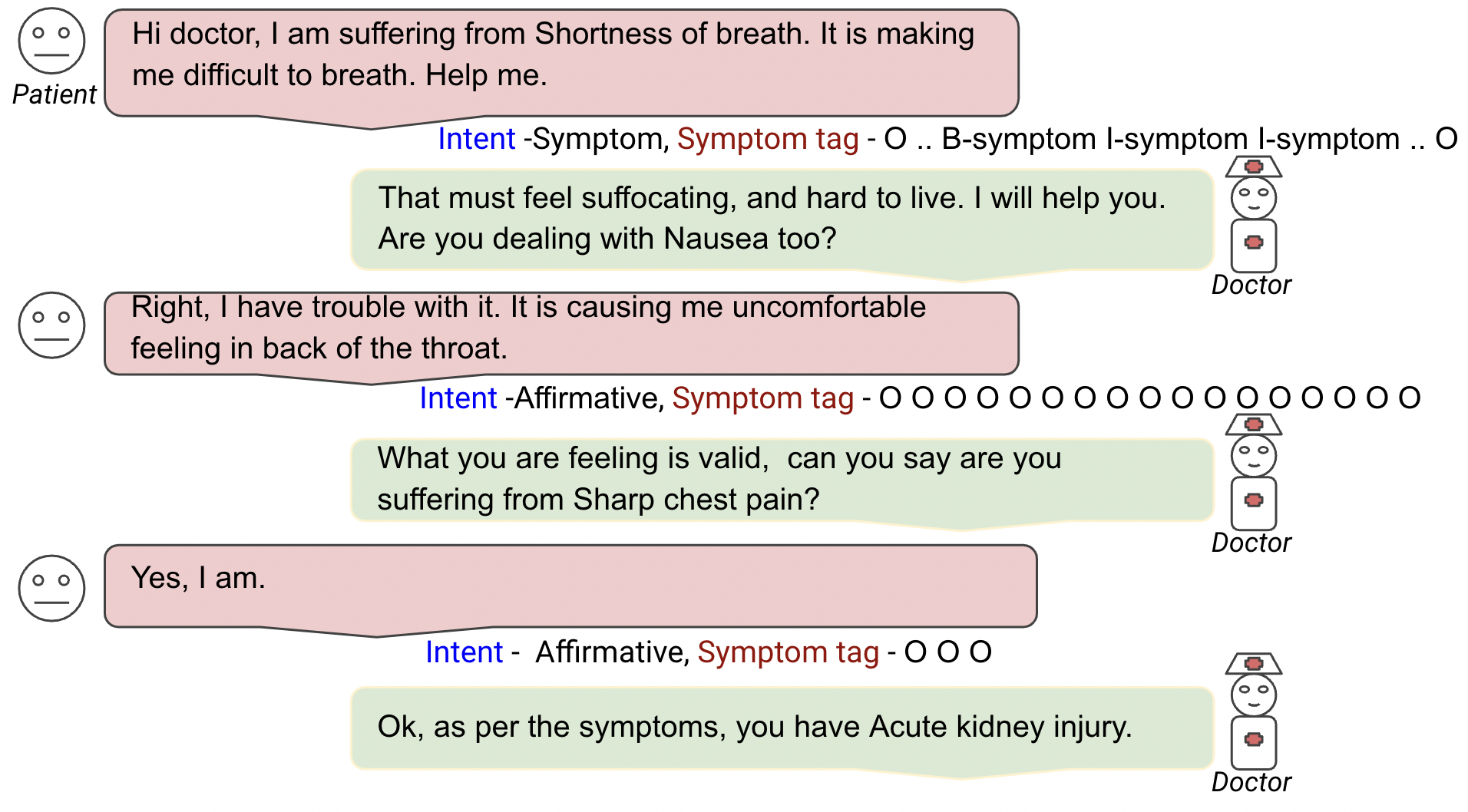}
    \vspace{-1em}
    \caption{A dialogue sample from the curated Empathical dataset. Conversation between patient and doctor having symptom and intent tagged. }
    \label{fig3}
\end{figure*}

 \hspace{-0.56cm}\textbf{Purpose of Intent and Symptom Information} Identifying intent and slot are two key tasks in NLU, which are vital for communicating with humans effectively. So, for building the NLU module, we have tagged intent and slot (here symptom) information for every utterance by the user (Figure \ref{fig3}).

\hspace{-0.56cm}\textbf{Role of Empathy} Patients' comfort and user satisfaction are of the utmost importance during doctor-patient consultations. This helps build trust between patient and doctor and increases patient compliance. Moreover, patients' recovery rate gets better when they connect with a doctor on common grounds, which boosts their mental well-being.

\section*{Methodology}

\begin{figure*}[hbt!]
    \centering
    \includegraphics[width=\linewidth]{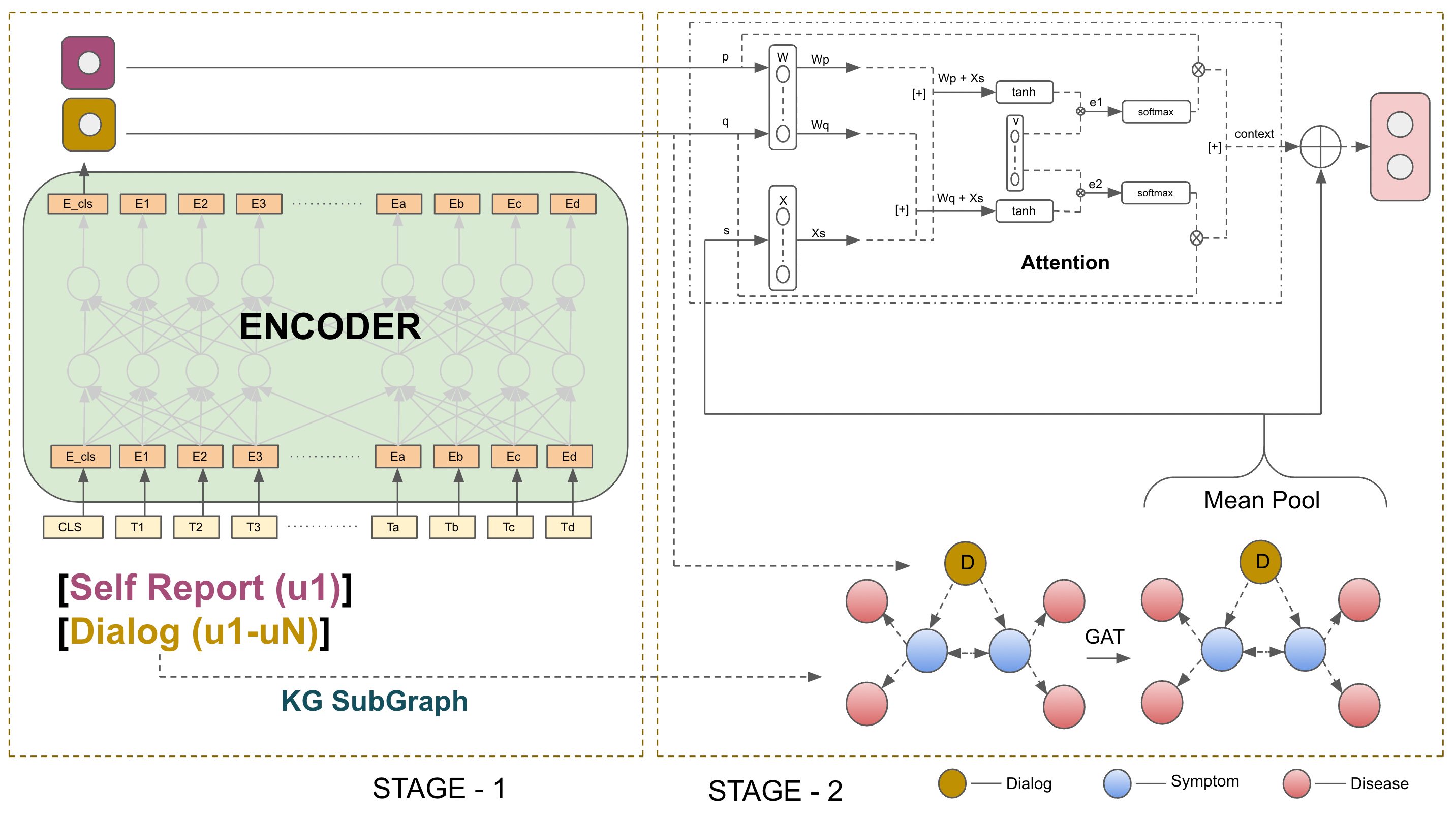}
    \vspace{-2em}
    \caption{KI-DDI: Self Report and Dialog are passed through the language model to obtain their embedding. The blue nodes are symptoms and Red nodes are diseases linked to symptoms. A joint Graph is formed by connecting the dialog node to all symptom nodes.}
    \label{KI-DDI}
\end{figure*}

We proposed a two-stage discourse-aware disease diagnosis framework; the two stages are (a) symptom investigation encoding and (b) external relevant knowledge infusion. The proposed architecture is illustrated in Figure \ref{KI-DDI}. The rationale behind the model is that for obtaining dialog and self-report embedding, we pass through the transformer encoder, i.e. SapBERT, and to diagnose the disease properly, we take help from external knowledge. To represent this external knowledge, we identify which diseases are more commonly linked with symptoms in the conversation and form a knowledge subgraph between symptoms and diseases. Then, to identify which symptom and its associated disease is more important in diagnosing disease, we form a joint graph between dialog embedding, symptoms, and diseases and apply Graph Attention Network. Finally, we use joint graph embedding to attend to dialog and self-report embedding to determine which is more critical for diagnosing, and then we diagnose the patient's disease. External knowledge helps aid clinicians by providing relevant information on which diseases are more closely linked with a particular symptom and providing knowledge expansion.

The model is comprised of three parts: \textbf{(i)} Symptom Investigation Encoding: Dialog and Self-Report Encoder, which generates the embedding for complete dialog between doctor and patient and also embeds the self-report given by the patient. Self-report signifies patients' chief complaints/major difficulties expressed by themselves. \textbf{(ii)} Knowledge Infusion: Knowledge Graph Extraction for extracting relevant sub-graphs from the knowledge graph to emphasize information relevant to the context. \textbf{(iii)} Disease Diagnosis Network. We have discussed and demonstrated the working principle of each module in the following sections.

\subsection*{Symptom Investigation Encoding}
Symptom investigation is the foundational and essential component of disease diagnosis. Patients first report their chief complaints; doctors conduct a thorough investigation and diagnose accordingly. Thus, encoding the investigation report efficiently is critical to the autonomous disease diagnosis model. Usually, doctors diagnose a disease based on the set of symptoms experienced by patients; however, they prioritize a few symptoms, particularly the patient's self-reported symptoms, in diagnosis. Thus, we segregate self-reported symptoms from the other extracted symptoms and infuse a weighted vector of this information into the diagnosis prediction model. To encode patient self-report and dialog, we have utilized SapBERT \cite{liu2020self} to capture the semantic meaning of patient-doctor utterances (Figure \ref{KI-DDI}). We have utilized $sr_{strat}$, $sr_{end}$ for denoting the self-reports start and end, respectively. Two more special tokens, $pat$, and $doc$ have been used to signify the starting position of patient and doctor utterances, respectively. We use SapBERT to get contextualized encoded representations ($S$ and $C$) from the vectors (Equations \ref{a} and \ref{b}).

\begin{equation}
 S = LM (\vert sr_{start}\vert SR \vert sr_{end} \vert)
\label{a}
\end{equation}

\begin{equation}
C = LM(\vert pat \vert Us_{1:t} \vert doc \vert  Do_{1:t})
\label{b}
\end{equation}
where $Us_i$ and $Do_i$ denote $i^{th}$ user utterance and doctor utterance, and $LM$ is the notion for the utilized language model (SapBERT). 

\subsection*{Knowledge Infusion} Clinical knowledge helps clinicians narrow the investigation space and use the information gathered efficiently during the diagnostic process. Thus, we aim to infuse the knowledge structure in the disease diagnosis framework.

\hspace{-0.48cm}\textbf{Knowledge Graph Construction} Here, we first created the knowledge graph (S-S-D) from the Empathical dataset, where symptoms (S) and diseases (D) are nodes. An edge between two nodes indicates their co-occurrence. The edges are weighted through the symptom frequency-inverse disease frequency (sf-idf) method \cite{ramos2003using} which involves applying the technique term frequency-inverse document frequency (TF-IDF) in symptom disease settings. Here, symptom frequency is equivalent to term frequency, and inverse disease frequency is equivalent to inverse document frequency. The edge weights between symptom-disease e(s, d, D) and symptom-symptom e($s_i$, $s_j$) are computed as follows:
\begin{equation}
   \small e(s, d, D) = sf(s, d) * idf(s, D)
\end{equation} 
\begin{equation}
    \small sf(s, d) = \frac{n_{sd}}{ \sum_{k} n_{kd}}  
\end{equation}
Here, $n_{sd}$ is the number of cases where symptom (s) has occurred with the disease, $d$. $k$ ranges in symptom space. The term $sf(s,d)$ represents the raw count of the co-occurrence of a symptom $s$ with disease $d$ divided by the co-occurrence of every symptom with disease $d$. 
\begin{equation}
    \small idf(s,D) = \log \frac{|D|}{|d: s \in disease_j|}  
\end{equation}  

Here $|D|$ signifies the total number of diseases. The term $idf(s,D)$ represents the logarithmic fraction of diseases containing the symptom $s$ obtained by dividing the total number of diseases by the number of diseases having symptom $s$ and then taking the logarithm of that quotient.

\begin{equation}
   \small e(s_i, s_j) = \frac{n(s_i, s_j)}{\sum_k n(s_i, s_k)}
\end{equation}

The term $e(s_i, s_j)$ represents the number of times symptoms $s_i$ and $s_j$ occur together, divided by the co-occurrence of symptom $s_i$ with all other symptoms. The intuition behind the symptom frequency-inverse disease frequency (sf-idf) is that the weight of symptom disease depends on the factor that if a symptom occurs with a particular disease and it also co-occurs with a large number of diseases, its inverse disease frequency will be close to zero (the denominator in “idf” will be closer to numerator) so the weight of that symptom and disease will be lower (since weight is product of symptom frequency(sf) and inverse disease frequency (idf)), indicating that the symptom is loosely associated with the disease. 
 If a symptom occurs with a particular disease and it also co-occurs with a very small number of diseases, then inverse disease frequency will be large (the denominator in “idf” will be much smaller than the numerator) so the weight of that symptom disease will be much higher, indicating that symptom is closely associated with that particular disease.

% where $n_{sd}$ is the number of cases where symptom (s) has occurred with the disease, $d$. $k$ ranges in symptom space, and $|D|$ signifies the total number of diseases. \\

\hspace{-0.56cm}\textbf{Knowledge Distillation}
While knowledge is crucial, focusing on relevant knowledge is more significant while solving a task. Thus, infusing the entire medical knowledge with the proposed disease diagnosis setup would be ineffective and may even deteriorate the performance. Thus, 
the proposed model extracts a subset of the knowledge graph depending on context (patients' symptoms) dynamically. It first extracts medical entities (signs and symptoms) from the conversation using joint BERT \cite{chen2019bert} language model and filters the knowledge graph considering the top $K$ associated diseases of the symptoms present in the conversation. We experimented with various $K$ values (1, 2, 3).

\begin{algorithm}[hbt!]
% \algsetup{linenosize=\tiny}
\scriptsize
\caption{\small \textbf{Discourse-aware Selective Filtering (DSF)}}
\label{DSF}
 \textbf{Initialization}: KG\_distill = \{\} \\
\textbf{Input}: Current Knowledge Graph - 
 KG\_distill, Complete Knowledge Graph - KG\_original, Dialog - dialog, Disease list - disease\_list, Symptoms extraction from text - SymptomExtractor.  \\
\textbf{Output}: Filtered Knowledge Graph (KG\_distill) 
\begin{algorithmic}[1] %[1] enables line numbers
\STATE symptom\_list = []
% \STATE KG$_{t+1}$ = KG$_t$
% \STATE Potential\_diseases (PD) =  $\Pi_{i=1}^{i=3}$ i$^{th}$-most\_associated\_disease(PSR) \hfill
\FOR{symptom in SymptomExtractor(dialog)}
\STATE symptom\_list.append(symptom)
%\STATE  a$_{PSR-d}$ = $\frac{n(PSR, d)}{|d|}$
% \STATE KG$_{t+1}$ = append(KG$_{t+1}$, triplet)
\ENDFOR
% \STATE symptom\_symptom\_weight = [], symptom\_disease\_weight = [] 
\FOR{i in len(symptom\_list)} 
\FOR{j in len(symptom\_list)}
\IF{i != j}
\STATE s\_i = symptom\_list[i], s\_j = symptom\_list[j]
\STATE symp\_symp\_weight\_1 = KG\_original[s\_i][s\_j]

\STATE symp\_symp\_weight\_2 = KG\_original[s\_j][s\_i]
\STATE KG\_distill.append(\{s\_i, s\_j, symp\_symp\_weight\_1\})
\STATE KG\_distill.append(\{s\_j, s\_i, symp\_symp\_weight\_2\})
\ENDIF
\ENDFOR
\ENDFOR

\FOR{i in len(symptom\_list)}
\STATE s\_i = symptom\_list[i]
\FOR{j in KG\_original[s\_i]}
\STATE disease\_count = 0
\IF{j in disease\_list}
\STATE  d\_j = j
\STATE symp\_dis\_weight = KG\_original[s\_i][d\_j]
\STATE KG\_distill.append(\{s\_i, d\_j, symp\_dis\_weight\})
\STATE disease\_count += 1
\IF{disease\_count == 3}
\STATE \textbf{break}
\ENDIF
\ENDIF
\ENDFOR
\ENDFOR

 \STATE \textbf{return} KG\_distill
 % \vspace{-0.5em}
\end{algorithmic}
% \vspace{-0.5em}
\end{algorithm}

\hspace{-0.56cm}\textbf{Graph Attention Network and Knowledge Infusion} 
We always prefer to analyze structured data, which helps us summarize effectively and take action path accordingly. Similar behavior has been observed for autonomous models, and thus, graph-based models are gaining huge popularity for developing models with a considerable amount of data \cite{zhang2020deep}. Motivated by efficacy, we build a graph attention (GAT) network over the relevant knowledge graph (S-D) and infuse it with context for disease diagnosis. In GAT, the vertex $i$ of l-th layer can be described by the following equation
\begin{equation}
    h_i^{(l)} = LeakyReLU(\sum_{j \in N_i} \alpha_{ij}W_hh_j^{(l-1)})
\end{equation}
% \[h_i^{l} = LeakyReLU(\sum_{j \in N_i} \alpha_{ij}W_hh_j^{l-1})\]

% \times{{d^{\prime}_1}}

where $N_i$ is the first hop neighbour of vertex i, $W_h$ $\in$ $\mathbb{R}^{d^{\prime}_1\times{{d_1}}}$ is trainable parameter. The attention weight $\alpha_{ij}$ for vertex $i$ is calculated as follows:

\begin{equation}
    \alpha_{ij} = \frac{exp(LeakyReLU(a^\mathsf{T}[W_hh_i||W_hh_j])}{\sum_{k \in N_i}exp(LeakyReLU(a^\mathsf{T}[W_hh_i||W_hh_k])}
\end{equation}
% \[\alpha_{ij} = \frac{exp(LeakyReLU(a^\mathsf{T}W_h[h_i||h_j])}{\sum_{k \in N_i}exp(LeakyReLU(a^\mathsf{T}W_h[h_i||h_k])}  \]

where a $\in$ $\mathbb{R}^{2d^{\prime}_1}$ is a trainable parameter. Here $||$ means concatenation. Finally, we obtain graph embedding by taking the mean pool of embedding of each vertex in JointGraph.  

\begin{equation}
    s = MeanPool(h_v^{(L)} | v \in Joint Graph) \in \mathbb{R}^{d_1}
\end{equation}

where $L$ denotes the last layer of GAT. $MeanPool$ is the average of node features across node dimensions. $JointGraph$ means graph obtained after adding dialog node to knowledge subgraph. The obtained global mean pool from the graph is passed to the attention layer.

\hspace{-0.47cm}\textbf{Attention Layer} 
In some cases, patient-reported data is crucial to understanding disease, while in other cases, symptoms extracted by physicians are crucial. We use additive attention  \cite{bahdanau2014neural} to compute attention. Here, the output of the GAT acts as a query, and self-report encoding and encoded dialog act as values. We take the weighted average of self-report encoding and dialog encoding and concatenate that with GAT output to finally pass it through the linear layer to perform disease classification. It can be expressed as follows:
\begin{equation}
    e_i = v^\mathsf{T}tanh(W_1h_i + W_2s)
\end{equation}
% \[ e_i = v^\mathsf{T}tanh(W_1h_i + W_2s) \] 

Here query s $\in$ $\mathbb{R}^{d_1}$ represents GAT output, i $\in$ $\{1,2\}$ where values $h_1$ $\in$ $\mathbb{R}^{d_2}$ means self report encoding and $h_2$ $\in$ $\mathbb{R}^{d_2}$ means dialog encoding.
Here $W_1$ $\in$ $\mathbb{R}^{d_3\times{d_2}}$ , $W_2$ $\in$ $\mathbb{R}^{d_3\times{d_1}}$ and $v$ $\in$ $\mathbb{R}^{d_3}$ are learnable parameters. Here $e_i \in \mathbb{R}$, e = [$e_1;e_2$] $\in \mathbb{R}^2$. The attention value $\alpha$ and final context is determined as follows:
% {\color{blue}Pls check if every term is defined}

\begin{equation}
    \alpha = softmax(e) \in \mathbb{R}^{2}
\end{equation}
% \[\alpha = softmax(e) \in \mathbb{R}^{2}\]

\begin{equation}
    context = \sum_{i = 1}^2 \alpha_ih_i \in \mathbb{R}^{d_2}
\end{equation}
% \[context = \sum_{i = 1}^2 \alpha_ih_i \in \mathbb{R}^{d_2}\]

Finally, the attended context is passed to the disease diagnosis network for disease prediction. 

\subsection*{Disease Diagnosis Network} 
We hypothesized that only patient self-report is not enough for disease diagnosis; we also need to consider doctor-patient interaction and additional medical knowledge for diagnosing patients effectively. Thus, our prediction network leverages all three components. We utilize self-report, doctor-patient interaction, and medical knowledge (joint graph) and pass the concatenation of attended vector (patient self-report, patient-doctor utterances, and knowledge graph) from the previous stage and joint graph embedding to a fully connected feed-forward network.  

\begin{equation}
    h_f = \sigma(W[s;context] + b)
\end{equation}

Here $\sigma$ is the softmax activation function. $W$ $\in$ 
$\mathbb{R}^{n \times (d_1 + d_2)}$ and $b \in \mathbb{R}^{n}$. $n$ is the number of diseases.

\begin{equation}
    \hat{y} =  argmax_i P(D_i \vert h_f) 
\end{equation}
where $i$ ranges over the set of diseases. We have utilized categorical cross entropy for calculating loss, which can be expressed as below.

\begin{equation}
L= -\sum_{i=1}^m\sum_{j = 1}^n y_j^{(i)}\log(\hat{y}_j^{(i)}) \in \mathbb{R} 
\end{equation}
where $m$ is the number of training examples, $n$ is the number of diseases. Here, $y_j^{(i)}$ is the ground truth label, and $\hat{y}_j^{(i)}$ is the predicted disease distribution label for $i^{th}$ dialogue. 

\subsection*{Experimental Setup}
We have used the curated Empathical dataset for training and evaluating the proposed model. We divided the dataset as follows: 70\% training, 10\% validation, and 20\% testing. We have utilized the PyTorch framework for implementing the proposed discourse-aware disease diagnosis model. We use SapBERT \cite{liu2020self} for encoding the dialog. In Table \ref{hyper}, we have listed the final values of hyperparameters. These values have been chosen through empirical experimentation using the validation dataset. The dataset we use is in English and created based on a benchmarked medical database SD Dataset\cite{zhong2022hierarchical}. The proposed model has been trained, validated, and tested with the dataset. The model works for English; however, it can be adapted to another language with minimal change, such as multi-lingual tokenizer incorporation. We use the BERT tokenizer, capable of processing slang words based on its pre-trained vocabulary, which includes a mix of formal and informal language from diverse sources. If a slang word is present in the vocabulary, BERT tokenizes it like any other word; otherwise, it may employ subword tokenization for out-of-vocabulary terms. Furthermore, the model's ability to handle slang depends on exposure to such terms during pre-training. While it excels with common slang, it may struggle with more niche or emerging expressions. Furthermore, the dataset and code are available at \url{https://github.com/NLP-RL/KI-DDI}.

\begin{table}[hbt!]
    \centering
    \scalebox{0.89}{
    \begin{tabular}{|l|l|l|}
     \hline
       \textbf{Hyperparameters}  & \textbf{Selected Values}  \\ 
       \hline
       Max sequence length  & 512 \\
       Batch Size & 16   \\ 
       GAT layers & 2  \\
       GAT hidden dim & 384\\
       GAT Attention heads & 3 \\
       GAT dropout & 0.5 \\
       Attention layer hidden dim 1 & 768 \\
       Attention layer hidden dim 2 & 384 \\
       Attention layer projection dim & 64 \\
       Optimizer & Adam \\
       Loss function & CrossEntropyLoss \\
       Learning rate & 1e-3 \\
       Epochs & 25 \\
       \hline
    \end{tabular}}
    \caption{Different hyperparameters and their values}
    \label{hyper}
\end{table}

\section*{Result and Discussion}
In order to comprehend the efficacy and limitations of the proposed model, we compared it with the following baselines and state-of-the-art models. The baselines and state-of-the-art models are as follows:
\vspace{-0.5em}
\begin{itemize}
    \item \textbf{BioLinkBert }\cite{yasunaga2022linkbert} - It is the pretraining method that uses links between different documents to train BERT. BioLinkBert is pretrained on PubMed articles with citation links on two self-supervised tasks masked language modeling and document relation prediction.
    \vspace{-0.5em}
    \item \textbf{KrissBert} \cite{zhang2022knowledge} - It trains PubMedBERT using entity list to generate self-supervised mention examples of biomedical entities and further it uses contrastive learning for training.
     \vspace{-0.5em}
    \item \textbf{KI-CD} \cite{tiwari2022knowledge} - It has a potential candidate (PCM) module which is based on Bayesian learning for symptom investigation. Also, it uses Hierarchical Reinforcement Learning for diagnosing disease.
     \vspace{-0.5em}
    \item \textbf{Coder} \cite{yuan2022coder} - It uses contrastive learning along with Unified Medical Language System (UMLS) knowledge graph to produce the embedding for medical terms.
     \vspace{-0.5em}
    \item \textbf{SapBert} \cite{liu2020self} - It trains the language model in a way that uses hard positive and hard negative samples to align synonymous biomedical entities. It uses a UMLS knowledge graph. 

\end{itemize}

\hspace{-0.60cm} \textbf{Evaluation Metrics} We utilize the most popular classification evaluation metrics, namely accuracy, F1-Score, and Jaccard similarity for evaluating the performance of different diagnosis models. The different metrics are defined as follows:

\begin{itemize}
    \item \textbf{Accuracy} It is defined as the number of correct predictions divided by the total number of predictions. It is represented as follows: \\
    \begin{equation}
    Accuracy = \frac{Number\ of\ correct\ predictions} {Total\ number\ of\ predictions}
    \end{equation}

    For a binary classification task, it can also be expressed as follows:
    \begin{equation}
    Accuracy = \frac{(TP + TN)}{(TP + TN + FP + FN)}
\end{equation}

Where TP - True Positive, TN - True Negative, FP - False Positive, FN - False Negative. 

True Positive (TP) - Number of examples predicted to be positive by the machine learning model and its label is actually positive. 

True Negative (TN) - Number of examples predicted to be negative by the machine learning model and its label is actually negative.

False Positive (FP) - Number of examples predicted to be positive by the machine learning model and its label is actually negative.

False Negative (FN) - Number of examples predicted to be negative by the machine learning model and its label is actually positive.

    \item \textbf{F1 score} is the harmonic mean of the precision and recall.
    \begin{equation}
        F1 = \frac{2 * (precision * recall)}{(precision + recall)}
    \end{equation}
    
    Precision  - It indicates the proportion of positive predictions that are actually correct. It is given as the ratio of True positive divided by the sum of True Positive and False Positive. 

    \begin{equation}
        Precision = \frac{(TP)}{(TP + FP)}
    \end{equation}

    Recall - It indicates the proportion of actual positives that are identified correctly. It is given as the ratio of True Positive divided by the sum of True Positive and False Negative.

    \begin{equation}
        Recall = \frac{(TP)}{(TP + FN)}
    \end{equation}

\item \textbf{Jaccard Similarity}  is defined as the size of the intersection divided by the size of the union of two label sets. It compares predicted labels to the ground truth labels for a sample. For ground truth label set “a” and predicted label set “b”, it is given as:

\begin{equation}
    Jaccard (a,b) = \frac{|a \cap b |}{|a \cup b|}
\end{equation}

\end{itemize}

All the reported values in the following tables are statistically significant, which are validated using the statistical t-test \cite{welch1947generalization} at a significant level of 5\%. The obtained performance by the joint BERT natural language understanding model for intent and symptom identification is provided in Table \ref{ISM}.  With the conducted experiments and performance comparison with the state-of-art/baseline models, the raised research questions (RQs) can be answered as follows.

\begin{table}[hbt!]
    \centering
    \scalebox{1}{
    \begin{tabular}{|l|l|l|}
     \hline
       \textbf{Task}  & \textbf{Accuracy(\%)} & \textbf{F1-Score}  \\ \hline
       Intent classification  & 95.49 & 0.9388 \\
       Symptom labeling & 92.04  & 0.9131 \\ \hline
    \end{tabular}}
    \caption{Performance of the joint intent and symptom module}
    \label{ISM}
\end{table}

\hspace{-0.56cm}\textbf{RQ1: Are self-reports from patients sufficient for accurate diagnosis?} Table \ref{T5} shows the efficacy of models that utilize only patient self-report for diagnosing a disease. The model that considers both self-reports and symptoms extracted by clinicians is way superior in terms of diagnostic accuracy. It firmly establishes the importance of the detailed symptom investigation conducted by clinicians. It is primarily due to the inadequacy of self-reports to accurately recognize patient diseases. It's also obvious because most of the time, we report symptoms that used to be common across several diseases, such as cold, cough, and fever. It shows the doctor needs to further investigate symptoms in addition to patient self-reports. Hence, the answer is no; we need further symptom investigation (in addition to patient self-report) to diagnose accurately.

\begin{table}[hbt!]
    \centering
    \begin{tabular}{|l|l|l|l|}
    \hline
      \textbf{Model}   & \textbf{Accuracy} & \textbf{F1-Score} & \textbf{Jaccard }  \\
      \hline
    %   KI-CD \cite{tiwari2022knowledge} & 57.84 & / &/ \\ 
       
       SRE + Linear & 23.80 & 0.183 & 0.122  \\ 
       Knowledge & 26.49 & 0.197 & 0.135\\
       SRE + Knowledge\_1 & 24.78 & 0.198 & 0.136\\
       SRE + Knowledge\_2 & 24.90 & 0.201 & 0.140\\
       SRE + Knowledge\_3 & 24.53 & 0.190 & 0.131\\
       
       \hline
    \end{tabular}
    \caption{Performance of model using Self Report with Linear and Knowledge. Here Knowledge\_2 means every symptom (blue node see  Figure \ref{KI-DDI}) has at most two diseases (red node) connected to it. }
    \label{T5}
\end{table}

\begin{figure}[hbt!]
    \centering
    \includegraphics[width=11cm, height=6.5cm]{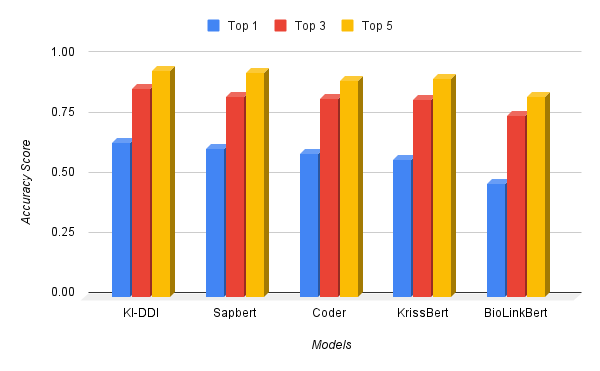}
    \vspace{-1.5em}
    \caption{Disease diagnosis accuracy of different models.}
    \vspace{-1em}
    \label{top_accuracy}
\end{figure}

\hspace{-0.56cm} \textbf{RQ2: How does the medical knowledge graph influence the disease diagnosis model's performance?} The performance obtained by the state-of-the-art model and our proposed knowledge-infused disease diagnosis models are reported in Table \ref{Tab4}. It shows that KI-DDI improved the performance of disease diagnosis by a margin of 2.57\% compared to the SapBERT \cite{liu2020self} model. Hence, we conclude that the knowledge graph is helpful in improving disease diagnosis accuracy. We also show the Top 1, 3, and 5 disease coverage accuracies for different models in Figure \ref{top_accuracy}.

\begin{table}[hbt!]
    \centering
    \begin{tabular}{|l|l|l|l|}
    \hline
      \textbf{Model}   & \textbf{Accuracy} & \textbf{F1-Score} & \textbf{Jaccard }  \\
      \hline
      BioLinkBERT \cite{yasunaga2022linkbert} & 47.25 & 0.4067 & 0.3129 \\
      KrissBERT \cite{zhang2022knowledge} & 57.14 & 0.4977 & 0.4091 \\
       KI-CD \cite{tiwari2022knowledge} & 57.84 & / &/ \\ 
       
       Coder \cite{yuan2022coder} & 59.70 & 0.5612 & 0.4630 \\
       SapBERT \cite{liu2020self} & 61.53 & 0.5801 & 0.4834 \\
       
       KI-DDI & \textbf{64.10}  & \textbf{0.6035} & \textbf{0.5099}\\
  
       \hline
    \end{tabular}
    \caption{Performance of the proposed KI-DDI model.  }
    \label{Tab4}
\end{table}

\hspace{-0.56cm}\textbf{RQ3: Does the mechanism of knowledge infusion impact the efficacy of disease diagnosis?} 
In addition to the knowledge being essential for autonomous models, its representation also matters. As humans, we always prefer to have information presented in a structured manner. With this motivation, 
we investigated the performance of different models having knowledge incorporated with different approaches. The obtained findings are reported in Table \ref{KG_R}. It demonstrates that the model that infuses medical knowledge using a graph structure outperformed the model that employs a linear structure by a significantly large margin of 4.76\%. Hence, the answer is yes, knowledge representation matters and the graph-based symptom-disease infusion is more effective in disease diagnosis than infusing it as a linear vector.

\begin{table}[hbt!]
    \centering
    \begin{tabular}{|l|l|l|l|}
    \hline
      \textbf{Model}   & \textbf{Accuracy (\%)} & \textbf{F1-Score} & \textbf{Jaccard}  \\
      \hline
    %   KI-CD \cite{tiwari2022knowledge} & 57.84 & / &/ \\ 
       
       SRE + Linear & 23.80 & 0.183 & 0.122  \\ 
       SRE + Knowledge  & \textbf{24.90} & \textbf{0.201} & \textbf{0.140}\\
       DE + Linear & 58.97 & 0.5306 & 0.4331  \\ 
       DE + Knowledge   & \textbf{63.73} \textbf{(4.76 $\uparrow$)} & \textbf{0.5752} \ & \textbf{0.4796} \\
       \hline
    \end{tabular}
    \caption{Performance comparison of adding Knowledge in Self Report and Dialog}
    \label{KG_R}
\end{table}

\hspace{-0.56cm}\textbf{Ablation Study} We have also conducted an ablation study to comprehend the importance of different components and concepts linked to the proposed model. The obtained findings are summarized in Table \ref{AS}. It leads to the following evidence and observations: \textbf{i.} We see that concatenating dialog encoding with knowledge graph embedding improves the performance. \textbf{ii.} We also observed that the behavior of constantly increasing knowledge width does not lead to superior performance, mainly because extraneous and large information sizes are included. \textbf{iii. } We see that using an attention mechanism between self-report encoding and dialog encoding leads to improvement. 

 \begin{table}[hbt!]
    \centering
    \begin{tabular}{|l|l|l|l|}
    \hline
      \textbf{Model}   & \textbf{Accuracy} & \textbf{F1-Score} & \textbf{Jaccard }  \\
      \hline
        Knowledge\_1 & 58.60 & 0.5353 & 0.4474\\ 
        Knowledge\_2 & \textbf{60.31} & \textbf{0.5479} & \textbf{0.4610}\\
        Knowledge\_3 & 58.48 & 0.5337 & 0.4446\\
        DE + Knowledge\_1 & 63.36 & \textbf{0.5909} & \textbf{0.4987}\\
        DE + Knowledge\_2 & 62.39
 & 0.5884 & 0.4903\\
        DE + Knowledge\_3 & \textbf{63.73} & 0.5752 & 0.4796\\
    %   SAPBERT with Linear &  58.24 & 0.519 & 0.425 \\
    %   SAPBERT with KG using GAT & 60.92 & 0.574 & 0.483\\
    %   SAPBERT with Linear and KG using GAT & 65.07 & 0.598 & 0.507\\
           KI-DDI\_1 & \textbf{64.10} & \textbf{0.6035} & \textbf{0.5099}  \\ 
      KI-DDI\_2 & 63.61 & 0.5969 & 0.5073\\ 
      KI-DDI\_3 & 63.24 & 0.5911 &  0.5007\\
    
        \hline
    \end{tabular}
    \caption{Result of the ablation study, which shows the efficacy of different components of the proposed model}
    \vspace{-1em}
    \label{AS}
\end{table}

\subsection*{Analysis}
\begin{figure}[hbt!]
    \centering
    \includegraphics[width=13cm, height=9cm]{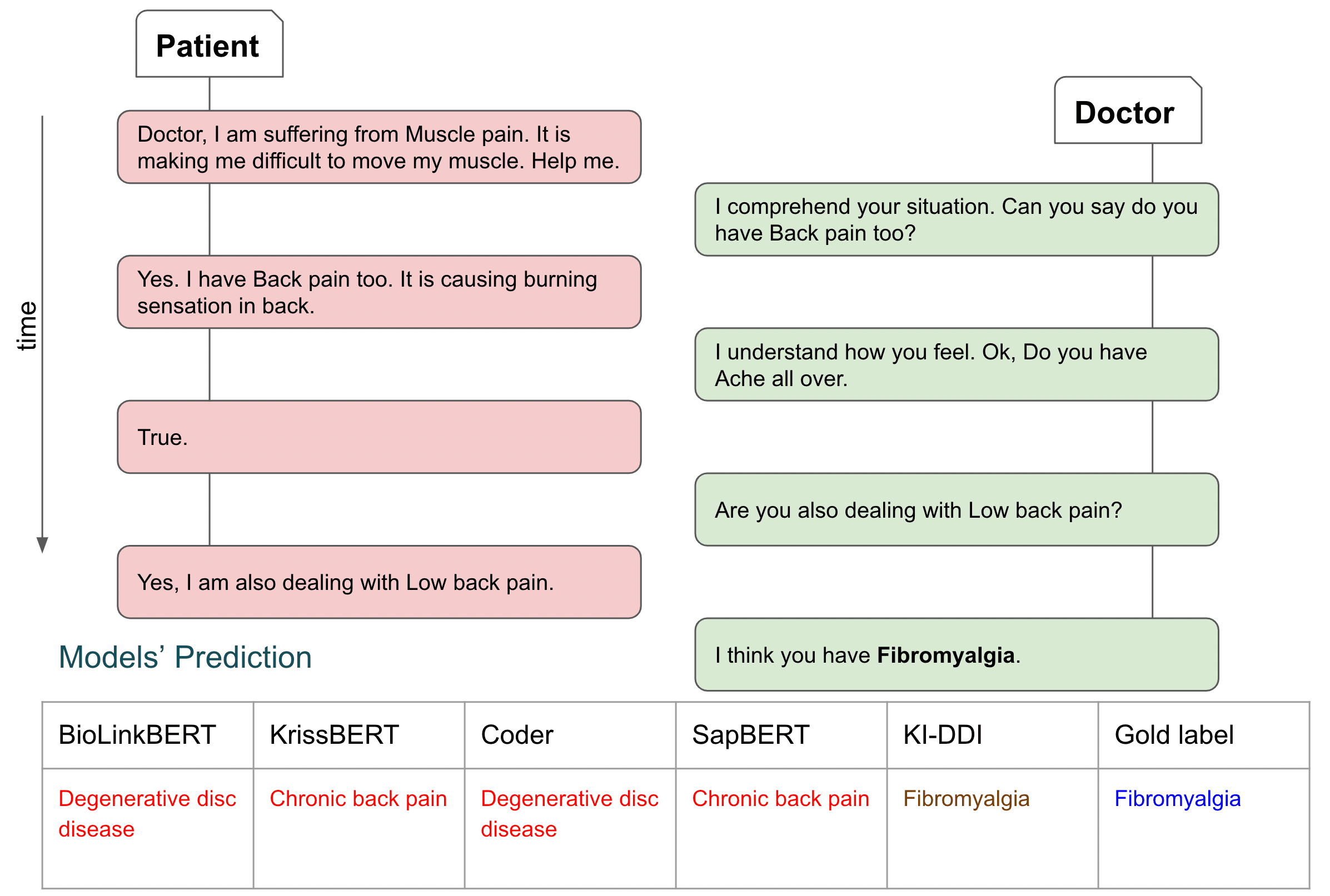}
    \vspace{-0.5em}
    \caption{Performance of KI-DDI and other models on a common test case.}
    \label{ki_vs_all}
\end{figure}

\begin{figure}[hbt!]
    \centering
    \includegraphics[width=6cm, height=4.3cm]{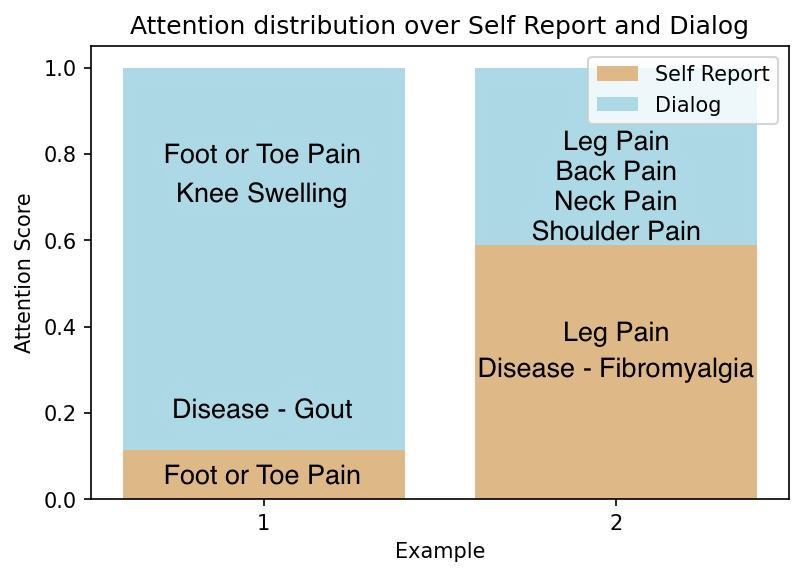}
    \vspace{-1em}
    \caption{Distribution of attention scores for two test examples.}
    \label{attention_scores}
\end{figure}

\begin{figure}[hbt!]
    \centering
    \includegraphics[width=9cm, height=5cm]{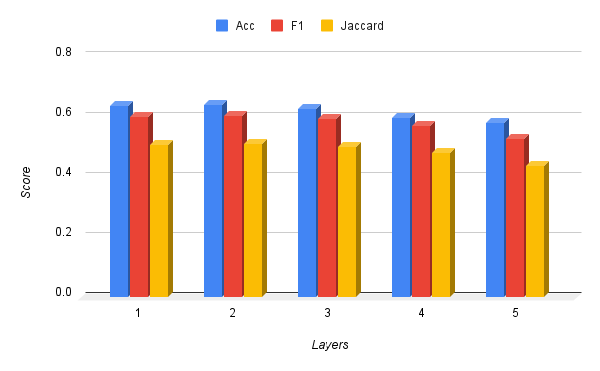}
    \vspace{-1.5em}
    \caption{Performance of KI-DDI model upon varying layers.}
    \label{layer_variation}
\end{figure}

The comprehensive analysis of the performances of different models led to the following major observations: \textbf{(i)} We analyze the performance of different models on common test cases and one such case study is shown in Figure \ref{ki_vs_all}. Our model correctly diagnosed the disease, while the other models misclassified the disease. This can be attributed to the knowledge infusion mechanism that the model is able to attend to symptoms that are more important for diagnosing the disease.  \textbf{(ii)} In order to exploit the structure of medical departments in healthcare systems, we also experimented with a hierarchical-based disease classifier. The first layer classifier triggers an appropriate medical department, and the activated disease classifier identifies the disease. The obtained results are reported in Table \ref{hierar}. \textbf{(iii)} In the case of hierarchical classification, we observed that the model identifies disease groups/medical departments quite adequately, but it gets confused among the diseases of the same medical group. \textbf{(iv)} We report the impact of variation of layers of GAT on the model's performance in Figure \ref{layer_variation}.  We find that upon increasing layers up to two model's performance increases then it starts decreasing.
\textbf{(v)} Sometimes patient self-report is vital to disease, whereas other times symptoms extracted by doctors are critical (Figure \ref{attention_scores}). We must thus take into account both and make a diagnosis that is appropriate to the situation rather than relying solely on one.

Our model also has some data biases like the majority of deep learning models; however, it is minimal while evaluating its impact. The model is trained on textual data, and some diseases have few examples, so our model is biased toward identifying diseases with many training examples. Also, many symptoms are expressed visually, and our model doesn’t integrate multi-modal input. Our model is trained on a single language corpus, i.e., English; its effectiveness is reduced in code-mixed scenarios. Our model has low diagnostic accuracy (64.10\%). Therefore, it can give inaccurate diagnoses and shouldn’t be used in real-world medical settings. But our model (KI-DDI) performs relatively better in the Top3 (86.8\%) and Top5 (94.01\%) accuracy in disease diagnosis.

 \begin{table}[hbt!]
    \centering
    \scalebox{0.9}{
    \begin{tabular}{|l|l|l|l|l|}
    \hline
      \textbf{Model}   & \textbf{Group Acc.} & \textbf{ Accuracy} & \textbf{ F1-Score} & \textbf{Jaccard }  \\
      \hline
    %   KI-CD \cite{tiwari2022knowledge} & 57.84 & / &/ \\ 
       
      KI-DDI\_1 & 79.24 & 58.97 & 0.555 & 0.457  \\ 
      KI-DDI\_2 & \textbf{81.19} & \textbf{62.27} & \textbf{0.583} & \textbf{0.487}\\ 
      KI-DDI\_3 & 80.70 & 60.07 & 0.565 & 0.467\\

      \hline
    \end{tabular}}
    \vspace{-0.5em}
    \caption{Hierarchical classification. Group Acc - Group Classification Accuracy, Acc - Disease classification within that group.}
    \vspace{-1em}
    \label{hierar}
\end{table}

 Table \ref{time_compare} shows that Bio Link Bert and KI-CD models take the highest train and inference time. BioLinkBert takes longer because of the bigger model size (having a total parameter count of 333 Million), and KI-CD takes longer train time because its architecture consists of 10 hierarchical models to train and has a longer inference time because it performs symptom investigation and disease diagnosis. In contrast, the remaining model only performs disease diagnosis. Models SapBert, KrissBert, and Coder take approximately the same train and inference time because of nearly the same parameter count (around 109 Million and 7 Thousand trainable parameters). KI-DDI takes longer because it is larger than SapBERT (as it involves SapBERT and Graph Attention Network), having (110 Million total parameters and 622 Thousand trainable parameters) but less time than BioLinkBERT because of its smaller model size.
 
\hspace{-0.6cm} \textit{Scalability} Our model KI-DDI utilizes a joint graph by incorporating a knowledge graph subgraph. It requires diseases associated with the symptoms. We experimented with each symptom associated with one to three diseases, and our model is scalable with more than three diseases associated with a symptom. 
 
\hspace{-0.56cm}\textit{Reliability} Our model achieves an accuracy of 64.10\% for disease diagnosis, which is low for diagnosing diseases in real-world settings. Hence, our model is not suitable for practical applications. But our model performs well in the Top3 (86.8\%) and Top5 (94.01\%) diagnosis accuracy. This shows that our model is getting confused to diagnose diseases linked common symptoms but works well in case of diagnosing diseases in the Top3 and Top5 settings.

\hspace{-0.56cm}\textit{Robustness} We have tested the robustness of our model concerning the number of layers and diseases linked with the symptoms and provided the results in Figure 7 and Table 8.

\begin{table}[]
\centering
\begin{tabular}{|l|l|l|}
\hline
\textbf{Model}       & \textbf{Training time (for 1 epoch)} & \textbf{Inference Time} \\ \hline
KI-DDI      & 22.26 sec                   & 5.00 sec       \\
BioLinkBert & 30.34 sec                   & 8.27 sec       \\

SapBert     & 9.55 sec                    & 2.57 sec       \\
KrissBert   & 7.91 sec                    & 2.52 sec       \\
Coder       & 9.35 sec                    & 2.52 sec       \\
KI-CD       & 30.14 sec                   & 9.44* sec          \\ \hline
\end{tabular}
    \vspace{-0.5em}
    \caption{Train and Inference time comparison of various models. Here, * means the model performs symptom investigation along with disease diagnosis.}
    \vspace{-1em}
    \label{time_compare}
\end{table}

\hspace{-0.56cm}\textbf{Limitations} While the proposed KI-DDI has demonstrated superior performance compared to baseline models, certain limitations have been observed. The key limitations are as follows: (i) The model has been trained and evaluated solely on a single-language corpus, specifically English. It exhibits reduced effectiveness when encountering code-mixed sentences. Therefore, an important avenue for future work is the incorporation of multilingual capabilities.  (ii) The model's performance across different diseases is influenced by the frequency of disease samples in the training data. Consequently, it may not perform well when there are very few samples available for certain diseases. Therefore, it is essential to integrate few-shot learning capabilities to address this limitation. (iii) Many symptoms are often conveyed through visual cues, but the model currently operates exclusively with text-based data. In future developments, we aim to integrate a multi-sensory input processing module into the diagnostic assistant.

\section*{Conclusion}
In this paper, we investigate the importance of knowledge infusion and doctor-driven symptom research in identifying patients' illnesses through dialogue. We presented a two-channel knowledge-infused, discourse-aware disease identification (KI-DDI) model that leverages external medical knowledge encoded through a context-aware filtered knowledge graph for identifying diseases accurately. We first developed a conversational disease diagnosis dataset in English, which is comprised of patient-doctor communication and annotated with semantic information (intent and symptom). The proposed model outperformed baselines and the existing state-of-the-art model significantly across all evaluation metrics. With the rigorous set of experiments conducted, the work evidences the paramount importance of (a) medical knowledge infusion, (b) doctor's collected symptoms (in addition to the patient's self-reported symptom), and (c) structured approaches for the knowledge representation. We note that the model's performance with respect to a specific disease is directly correlated with the quantity of samples available for that disease in the dataset. To mitigate this effect and ensure effective performance even for diseases with limited samples, the inclusion of a few-shot learning module could be considered. When we consult with doctors, we often report and describe our health conditions with visual aids. Moreover, many people are unacquainted with several symptoms and medical terms. Thus, we would like to extend the work by investigating the role of multi-modality in symptom investigation \& diagnosis and building a multimodal diagnosis dialogue system.

% \bibliography{sample}

\section*{Author contributions statement}

Mohit Tomar (M.T): Conceptualization, Data analysis, Experimentation, Validation, Analysis, Investigation, Visualization, and Writing; Abhisek Tiwari (A.T): Conceptualization, Data analysis, Experimentation, Validation, Analysis, Investigation, Visualization, and Writing; Sriparna Saha (S.S): Conceptualization, Data analysis, Validation, Analysis, Visualization, and Writing. All authors reviewed the manuscript.

\section*{Additional information}

\textbf{Ethical Consideration} While creating the dataset, we followed guidelines aligned with medical research's legal, ethical, and regulatory standards. We utilized a benchmarked dataset named SD (Zhong, Cheng, et al., 2022) to construct a conversational corpus. It's important to note that the dataset includes samples with user consent. With this in mind, we have not added or removed any entity in a conversation corresponding to the reported dialogues in the benchmark SD dataset. Also, the curated dataset does not disclose users' personal information. Hence, we ensure that the Empathical dataset and each step of its formation do not violate ethical and clinical principles. We have also obtained approval from our institute's ethical committee, IIT Patna, to employ the dataset and carry out the research (IITP/EC/2022-23/07). Please note that the research does not involve any human beings or living entities.

\hspace{-0.56cm}\textbf{Informed consent and Privacy:} We utilized a benchmarked dataset named SD (Zhong, Cheng, et al., 2022) to construct a conversational corpus. It's important to note that the dataset includes samples with user consent. They do not include any personal patient information, such as names, ages, or genders. Instead, they solely contain information about symptoms discussed during conversations with doctors and the identified diseases by the doctors.

\hspace{-0.56cm}\textbf{Societal ramifications:} Over the last five years, numerous surveys and reports have consistently highlighted an imbalanced doctor-to-population ratio. These findings strongly advocate for addressing the concerning statistics by augmenting the healthcare workforce and optimizing their time more effectively. With the objective of aiding doctors and streamlining early diagnosis, the suggested automated disease diagnosis assistant plays a pivotal role in assisting healthcare professionals in precisely identifying illnesses. The research delves into the impact of knowledge infusion on disease identification through doctor-patient conversations. Rigorous experiments and human analyses across diverse algorithms underscore the substantial influence of knowledge infusion in deducing diseases. 

\hspace{-0.56cm}\textbf{Reproducibility:} We have used the curated Empathical dataset for training and evaluating the proposed model. We divided the dataset as follows: 70\% training, 10\% validation, and 20\% testing. We have utilized the PyTorch framework for implementing the proposed discourse-aware disease diagnosis model. We use SapBERT (Liu et al., 2020) for encoding the dialog. In Table below, we have listed the final values of hyperparameters. These values have been chosen through empirical experimentation using the validation dataset. The dataset we use is in English and created based on a benchmarked medical database SD Dataset. The proposed model has been trained, validated, and tested with the dataset. The model works for English; however, it can be adapted to another language with minimal change, such as multi-lingual tokenizer incorporation. We have provided details of our experimental setup, including hyperparameter values and evaluation metrics, and made our code available (https://github.com/NLP-RL/KI-DDI).

\hspace{-0.56cm}\textbf{Accession codes}  We have made a GitHub repository that contains the curated conversational dataset and the experimental setup (code). The dataset and code are available at \url{https://github.com/NLP-RL/KI-DDI}.

\hspace{-0.56cm}\textbf{Competing interests} The authors declare that they have no competing interests. The corresponding author is responsible for submitting a \href{http://www.nature.com/srep/policies/index.html#competing}{competing interests statement} on behalf of all authors of the paper. This statement must be included in the submitted article file.

\end{document}